\documentclass[pdflatex,sn-aps]{sn-jnl}

\usepackage{graphicx}%
\usepackage{multirow}%
\usepackage{amsmath,amssymb,amsfonts}%
\usepackage{amsthm}%
\usepackage{mathrsfs}%
\usepackage[title]{appendix}%
\usepackage{xcolor}%
\usepackage{textcomp}%
\usepackage{manyfoot}%
\usepackage{booktabs}%
\usepackage{algorithm}%
\usepackage{algorithmicx}%
\usepackage{algpseudocode}%
\usepackage{listings}%
\usepackage{lineno}
\usepackage{makecell}
\usepackage[utf8]{inputenc}

\theoremstyle{thmstyleone}%
\theoremstyle{thmstyletwo}%

\theoremstyle{thmstylethree}%

\newcommand{\OceanDA}{ADAF-Ocean}

\begin{document}

\title[Article Title]{Advancing Ocean State Estimation with efficient and scalable AI}



\author[1]{\fnm{Yanfei} \sur{Xiang}}



\author[1]{\fnm{Yuan} \sur{Gao}}

\author[1]{\fnm{Hao} \sur{Wu}}

\author[1]{\fnm{Quan} \sur{Zhang}}

\author[1]{\fnm{Ruiqi} \sur{Shu}}

\author[1]{\fnm{Xiao} \sur{Zhou}}

\author*[1]{\fnm{Xiaomeng} \sur{Huang}}\email{hxm@tsinghua.edu.cn}


\affil[1]{\orgdiv{Department of Earth System Science, Ministry of Education Key Laboratory for Earth System Modeling, Institute for Global Change Studies, Tsinghua University}, \orgaddress{\city{Beijing}, \postcode{100084} \country{China}}}




\abstract{
Accurate and efficient global ocean state estimation remains a grand challenge for Earth system science, hindered by the dual bottlenecks of computational scalability and degraded data fidelity in traditional data assimilation (DA) and deep learning (DL) approaches. Here we present an AI-driven Data Assimilation Framework for Ocean (\OceanDA) that directly assimilates multi-source and multi-scale observations, ranging from sparse in-situ measurements to 4 km satellite swaths, without any interpolation or data thinning. Inspired by Neural Processes, \OceanDA~learns a continuous mapping from heterogeneous inputs to ocean states, preserving native data fidelity. Through AI-driven super-resolution, it reconstructs 0.25$^\circ$ mesoscale dynamics from coarse 1$^\circ$ fields, which ensures both efficiency and scalability, with just 3.7\% more parameters than the 1$^\circ$ configuration. When coupled with a DL forecasting system, \OceanDA~extends global forecast skill by up to 20 days compared to baselines without assimilation. This framework establishes a computationally viable and scientifically rigorous pathway toward real-time, high-resolution Earth system monitoring.
}





\maketitle

\section{Introduction}\label{sec:introduction}

The ocean is central to the Earth system, fundamentally regulating climate, driving global resource distribution, and sustaining ecosystems~\cite{flato2014, palmer2018, khatiwala2009}. 
Accurate modeling and forecasting of the global ocean are crucial for understanding the Earth system~\cite{Stockdale1998, Kirtman2012, Caldeira_2005}. However, the predictive capabilities of numerical ocean models are limited by initial condition uncertainties and the significant computational challenge posed by the rapidly increasing volume of observational data from multiple sources~\cite{fox2019, Hurlburt2009, Cui2025}.

Data assimilation (DA) is the established method to mitigate these limitations by optimally combining  observations with models~\cite{Jacobs2021, Huang2019, asch2016}. However, classical DA methods are grappling with a critical scaling crisis.
Methods such as four-dimensional variational (4DVar)~\cite{Lorenc1986, Bannister2017} and the ensemble Kalman filter (EnKF)~\cite{evensen2003, Carrassi2018} face computation costs that scale polynomially with resolution, making high-resolution global applications computationally infeasible.
Furthermore, these methods often rely on simplified physical assumptions, such as linearity or Gaussian error statistics, which fail to capture the complexities of strong non-linear dynamics inherent in the ocean system~\cite{Evensen1994Sequential}.
Another major challenge is integrating diverse observational data. Ocean data is highly heterogeneous, ranging from sparse in-situ profiles to high-resolution satellite swaths.
To cope with the data volume, current practice often uses observation thinning or averaging~\cite{Zhou2024}. However, it irreversibly sacrifices valuable fine-scale information. As a result, increasing data availability does not lead to proportional model improvement.

Recent advances in deep learning (DL) offer a promising avenue to overcome the limitations of traditional DA, excelling at capturing non-linear dynamics and accelerating computations. DL has already demonstrated success in enhancing DA components, such as observation operators~\cite{frerix2021variational}, dynamic error covariance estimation~\cite{cocucci_model_2021, Melinc_2024}, and online model error correction~\cite{Farchi2021Using}. Indeed, several DL frameworks show potential to entirely replace traditional DA methods in atmospheric and oceanic applications~\cite{huang2024diffda, Xiang2025, chen2024, ham2024, Xu2025}.
However, most existing DL-based data assimilation (DL-DA)  frameworks suffer from a critical architectural flaw: their reliance on vision-based DL models, including Convolutional Neural Networks (CNNs)~\cite{khan2020survey} and Vision Transformer (ViT)~\cite{dosovitskiy2021}, necessitates extensive preprocessing steps that often degrade data fidelity. Specifically, this preprocessing typically involves interpolating or downsampling all observations onto a uniform, regular grid~\cite{Xiang2025, Xu2025, chen2024}, irreversibly discarding high-frequency information and limiting the utility of high-resolution sensors.
Moreover, the substantial graphics processing unit (GPU) memory demands for training these architectures severely constrain their scalability and real-time applicability for high-dimensional DA tasks.

To address these challenges, we introduce \OceanDA~(AI-Driven Data Assimilation Framework for the Ocean). 
This novel DL-based framework fundamentally redefines the DA paradigm by overcoming the dual bottlenecks of computational scalability and degraded data fidelity in traditional DA and existing DL approaches.
Our core innovation lies in the ability of the framework to operate directly on raw, heterogeneous, and multi-scale observational data without any prior interpolation or data-degrading preprocessing.
Our Neural Process-inspired encoder-decoder architecture maximizes information extraction by handling diverse inputs—from sparse in-situ points to 4 km satellite swaths—at their native resolutions. The main contributions of this study are as follows:
\begin{itemize}
    \item \textbf{Maximized Data Fidelity}: A novel encoder-decoder architecture that circumvents data-degrading preprocessing, maximizing observations utilization and preserving data integrity.
    \item \textbf{AI-Driven Super-Resolution}: \OceanDA~physically reconstructs high-resolution (0.25$^\circ$) mesoscale dynamics even with a coarse 1$^\circ$ background field. This capability is achieved without increasing the complexity of model, requiring only 28 million parameters for 0.25$^\circ$ DA, a minimal increase from 27 million for 1$^\circ$ DA.
    \item  \textbf{Enhanced Forecast Skill}: Coupled with a DL forecast model, the analysis fields produced by \OceanDA~provide superior initial conditions, significantly improving forecast skill for up to 20 days.
\end{itemize}
\OceanDA~provides a computationally efficient, scientifically robust, and scalable approach to address the DA crisis. 
It enables the evaluation of the contributions of diverse observation types through an effective, data-driven methodology, facilitating the optimization of global ocean monitoring networks.
By maximizing the utility of observational data, this framework represents a critical step toward real-time, high-resolution Earth system monitoring and prediction.

\section{Results}\label{sec:results}

\subsection{Overall performance of analysis}

\begin{figure}
    \centering
    \includegraphics[width=1\linewidth]{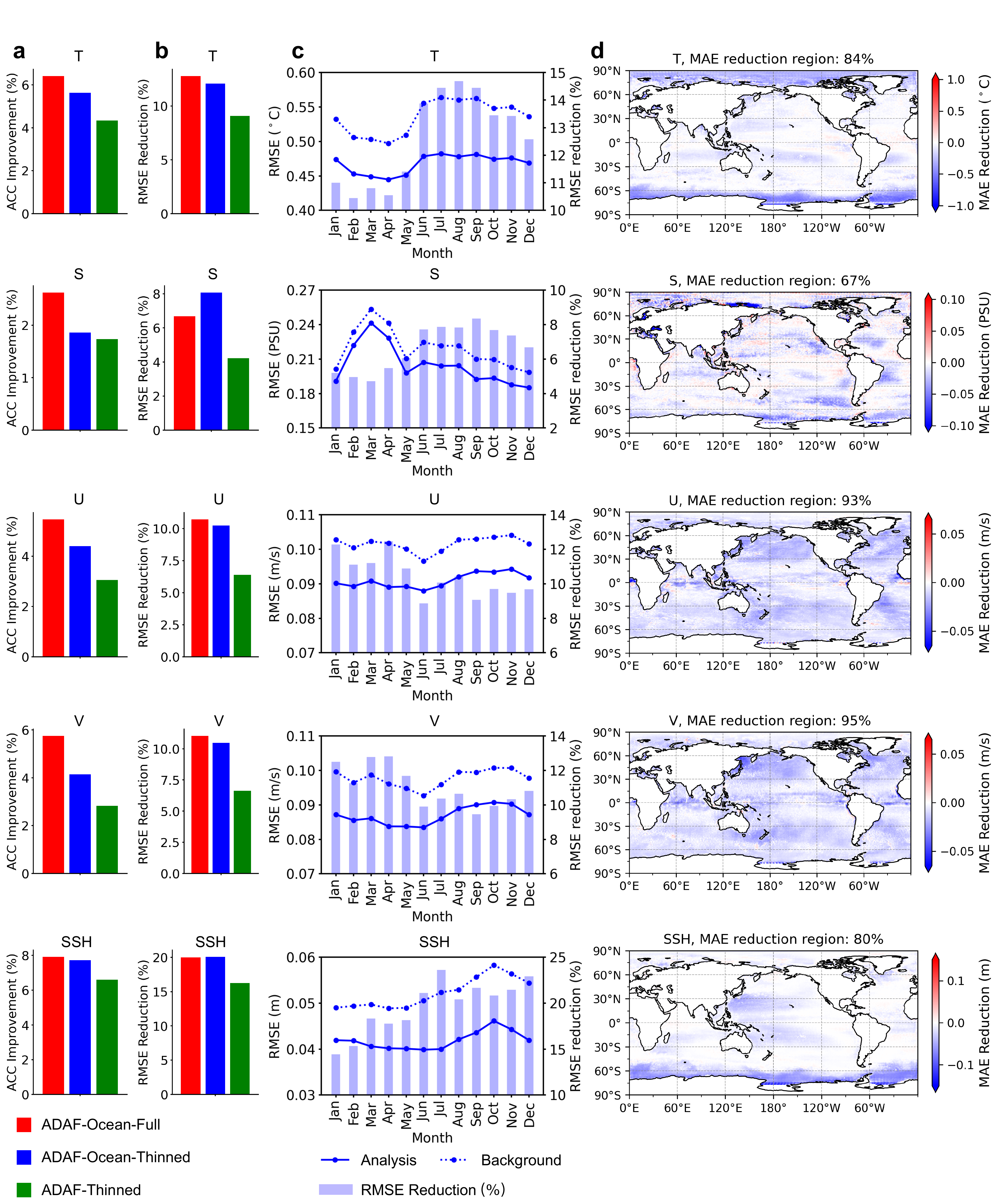}
    \caption{\textbf{Overall performance of analysis for 5 surface ocean variables.} \textbf{a, b} Comparison of \OceanDA-Full, \OceanDA-Thinned, and ADAF-Thinned  analysis in terms of ACC improvement and RMSE reduction. \textbf{c} Monthly latitude-weighted RMSE trends for the background and analysis produced by \OceanDA-Full, along with RMSE reduction for the analysis. \textbf{d} Spatial MAE reduction ($\text{MAE}_{\text{analysis}} - \text{MAE}_{\text{background}}$). Negative values (blue) indicate a reduction in MAE achieved by \OceanDA.}
    \label{fig:DA_metrics_with_baselines}
\end{figure}

To illustrate the ability of \OceanDA~in fully assimilating diverse observational data sources, six types of observations, including satellite and in-situ measurements that are irregularly and sparsely distributed, were utilized (refer to Section~\ref{sec:datasets} and Table~\ref{tab:observations}). The model was trained to reconstruct global surface ocean states using 3-day forecast fields from the DL ocean forecasting model Triton~\cite{Wu2025Triton} as the background, with GLORYS reanalysis serving as the target. Data from 2018 were used for training, 2019 for validation, and 2020 for testing (details in Section~\ref{sec:model_training}). The assimilation was performed daily.

Three experimental configurations were designed to evaluate the performance of \OceanDA: \OceanDA-Full, \OceanDA-Thinned, and ADAF-Thinned. In \OceanDA-Full, we assimilated observations at their original spatial resolutions, leveraging the ability of \OceanDA~to effectively utilize multi-source and multi-resolution data. In \OceanDA-Thinned, assimilated observations were preprocessed to a unified $1^\circ$ resolution. For a robust baseline comparison, ADAF-Thinned employed the existing deep learning framework ADAF (An Artificial Intelligence Data Assimilation Framework)~\cite{Xiang2025} to assimilate the same $1^\circ$ resolution data. The \OceanDA-Full configuration highlights the benefits of assimilating high-resolution observations, while \OceanDA-Thinned demonstrates the superior capability of \OceanDA in handling multi-source data compared to the established DL-DA method, ADAF-Thinned.

The analysis results (Fig.~\ref{fig:DA_metrics_with_baselines}) were evaluated against the GLORYS reanalysis reference, providing a comprehensive assessment of the effectiveness of \OceanDA~in assimilating multi-source, multi-resolution, and irregularly distributed observations.
The \OceanDA-Full configuration, which assimilates observations at their original fidelity (up to 4 km) by circumventing data-degrading preprocessing, consistently achieves the highest Anomaly Correlation Coefficient (ACC) improvement and superior Root Mean Square Error (RMSE) reduction across all surface variables (Fig.~\ref{fig:DA_metrics_with_baselines}{a, b}). 
ACC improvement is quantified as $(\text{ACC}_\text{analysis} - \text{ACC}_\text{background}) / \text{ACC}_\text{background}$, and RMSE reduction is quantified as $(\text{RMSE}_\text{background} - \text{RMSE}_\text{analysis}) / \text{RMSE}_\text{background}$.
This superior performance of \OceanDA-Full over the \OceanDA-Thinned and ADAF-Thinned baselines stems directly from the intrinsic capability of \OceanDA~to fully extract multi-scale observational information. This capability allows the model to leverage the rich, high-frequency details preserved in the original data, a feature unavailable to methods relying on thinned or homogenized inputs. 
Notably, the ACC improvements for T and SSH ($>$6\%) are significantly larger compared to the other configurations. A distinct result is observed for S: while \OceanDA-Full shows the highest ACC improvement (indicating superior spatial pattern correlation), its absolute RMSE reduction for S is comparatively smaller. This discrepancy arises because \OceanDA-Full effectively enhances the spatial correlation of the salinity field with the reference, yet simultaneously retains localized variability, which slightly limits the overall reduction in absolute errors.

The monthly averaged latitude-weighted RMSE for \OceanDA-Full is consistently lower than the background across all variables (Fig.~\ref{fig:DA_metrics_with_baselines}{c}). SSH achieves the largest RMSE reduction ($\sim$18\%), while S shows the smallest ($\sim$5\%). 
Significant seasonal patterns are observed. 
Improvements for T, S, and SSH peak from July to December, suggesting better constraint of density-driven variables due to denser satellite thermal data (SST) during summer and autumn, while velocity components (U and V) show maximum improvement from January to May, indicating optimal effectiveness when wind forcing (SSW) is typically strongest.
The monthly averaged latitude-weighted ACC is presented in Supplementary Fig.~\ref{appfig:monthly_acc}, showing improvements of approximately 6\% for T, 2.5\% for S, 5\% for U, 6\% for V, and 7\% for SSH compared to the background.
Additionally, the spatial distribution of MAE difference ($\text{MAE}_\text{analysis} - \text{MAE}_\text{background}$) (Fig.~\ref{fig:DA_metrics_with_baselines}{d}) was evaluated. The percentage of grid points showing MAE reduction reaches 85\%, 68\%, 93\%, 95\%, and 80\% for T, S, U, V, and SSH, respectively. 
For T, the most significant improvements are observed in the Arctic and Antarctic regions. SSH shows notable enhancements in the Kuroshio region, a highly energetic western boundary current, while U and V demonstrate improvements exceeding 90\% across the global ocean, with the greatest gains in the North Pacific where dynamic constraints are critical.

We further investigate the contributions of observations (Supplementary Fig.~\ref{appfig:obs_contrib}). The results highlight the critical role of satellite-based observations—particularly SST, SSW, and SSS—in enhancing the accuracy of ocean state estimation. 
To account for observational uncertainties, we assessed the sensitivity of state estimation to satellite observations under noise perturbations. The findings reveal that surface ocean state estimation is particularly sensitive to satellite SLA, with more details provided in Supplementary Section~\ref{appsec:uncertainty_sensitivity_analysis}.

\subsection{AI-Driven Super-Resolution Reconstruction}

\begin{figure}[h]
    \centering
    \includegraphics[width=0.95\linewidth]{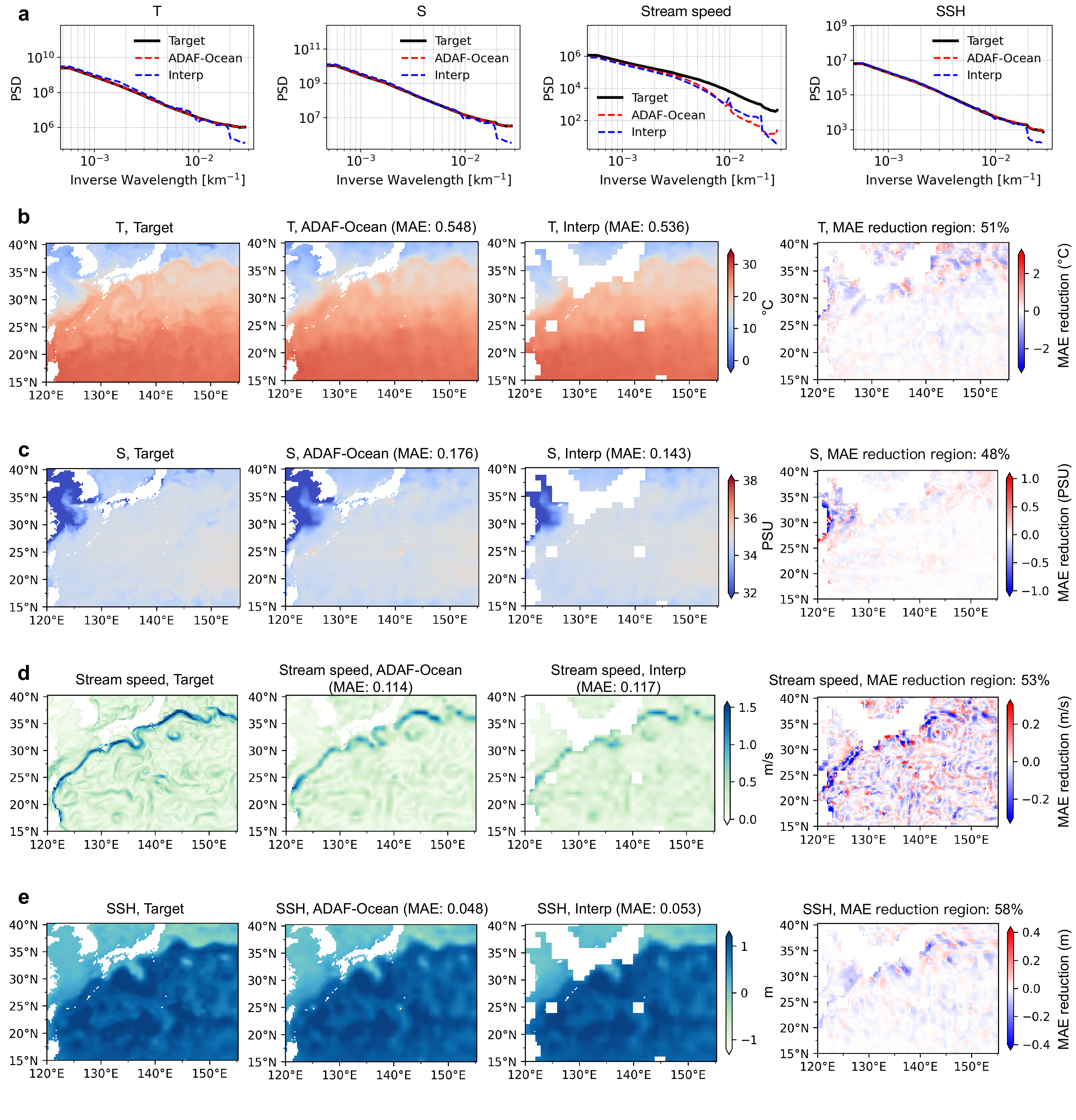}
    \caption{\textbf{AI-Driven Super-Resolution Reconstruction and Physical Fidelity.} The figure demonstrates the ability of \OceanDA~to reconstruct fine-scale features compared to linear interpolation (Interp). \textbf{a} Power Spectral Density (PSD) analysis from a representative test day across the global domain for T, S, Stream speed ($\sqrt{U^2 + V^2}$), and SSH, showing the superior capture of high-wavenumber (small-scale) features by \OceanDA. \textbf{b-e} Case studies in the highly dynamic Kuroshio region, showing the target from GLORYS reanalysis, $0.25^\circ$ analysis produced by \OceanDA~and Interp, and the resulting MAE reduction ($\text{MAE}_{\text{\OceanDA}} - \text{MAE}_{\text{Interp}}$). Negative values (blue) indicate lower MAE achieved by \OceanDA~compared to Interp.}
    \label{fig:high_resolution_DA}
\end{figure}

High-resolution ocean modeling significantly enhances ocean state fidelity, necessitating effective high-resolution data assimilation (HR-DA)~\cite{Smith2000, qiang_2024, chassi2021}.
To train this capability, we utilized the high-fidelity $0.25^\circ$ GLORYS reanalysis as the ground-truth target. We then demonstrate that \OceanDA~achieves AI-Driven Super-Resolution Reconstruction, generating $0.25^\circ$ analysis fields from a coarse $1^\circ$ background.
This is realized because our DL model learns a continuous DA operator that directly predicts the analysis increment, representing high-frequency error correction at the targeted $0.25^\circ$ resolution.
This high-fidelity reconstruction is achieved with minimal computational overhead, utilizing only 28 million parameters—a minimal increase from the 27 million required for $1^\circ$ DA—a significant contrast to the computational scaling typically required for traditional DA methods.

Crucially, the Power Spectral Density (PSD) analysis (Fig.~\ref{fig:high_resolution_DA}{a}) provides evidence of the physical reconstruction capability of \OceanDA. The PSD curve of \OceanDA~aligns  closer to the ground-truth GLORYS spectrum in the high-wavenumber regime (small scales) compared to the sharply attenuated PSD of Interp. This convergence confirms that \OceanDA~effectively captures and reconstructs the energy cascade associated with mesoscale and sub-mesoscale eddies, which are physical features that are entirely lost or misrepresented by linear interpolation.
To validate the ability of \OceanDA~in fine-scale reconstruction, we compared the performance of \OceanDA~against statistical linear interpolation (Interp) applied to a $1^\circ$ DA analysis. Globally, \OceanDA~consistently achieves superior performance with lower MAE for most variables compared to Interp of the $1^\circ$ analysis, confirming its reconstruction ability (Supplementary Fig.~\ref{appfig:high_resolution_DA_mae}).

Focusing on the highly dynamic Kuroshio region (Fig.~\ref{fig:high_resolution_DA}{b-e}), \OceanDA~captures finer-scale oceanic structures and achieves significant error reduction: the MAE for SSH is reduced to 0.048 and stream speed ($\sqrt{U^2 + V^2}$) to 0.114, substantially outperforming Interp.
This high-fidelity reconstruction is evident in the well-defined mesoscale eddies in the Stream Speed field (Fig.~\ref{fig:high_resolution_DA}{d}) and the SSH field (Fig.~\ref{fig:high_resolution_DA}{e}), which are strongly attenuated by Interp.
While \OceanDA~exhibits a slightly higher MAE for T and S, this is because the Interp method cannot produce valid values near complex land-sea boundaries.
These regions inherently exhibit high local variability in their T and S fields. The ability of \OceanDA~to reconstruct details in these challenging areas, rather than omitting them, results in a higher localized MAE compared to Interp.
Within this Kuroshio region, the percentage of grid points showing MAE reduction is 51\% for T, 48\% for S, 53\% for stream speed, and 58\% for SSH.

\subsection{Forecast verification}

\begin{figure}
    \centering
    \includegraphics[width=1\linewidth]{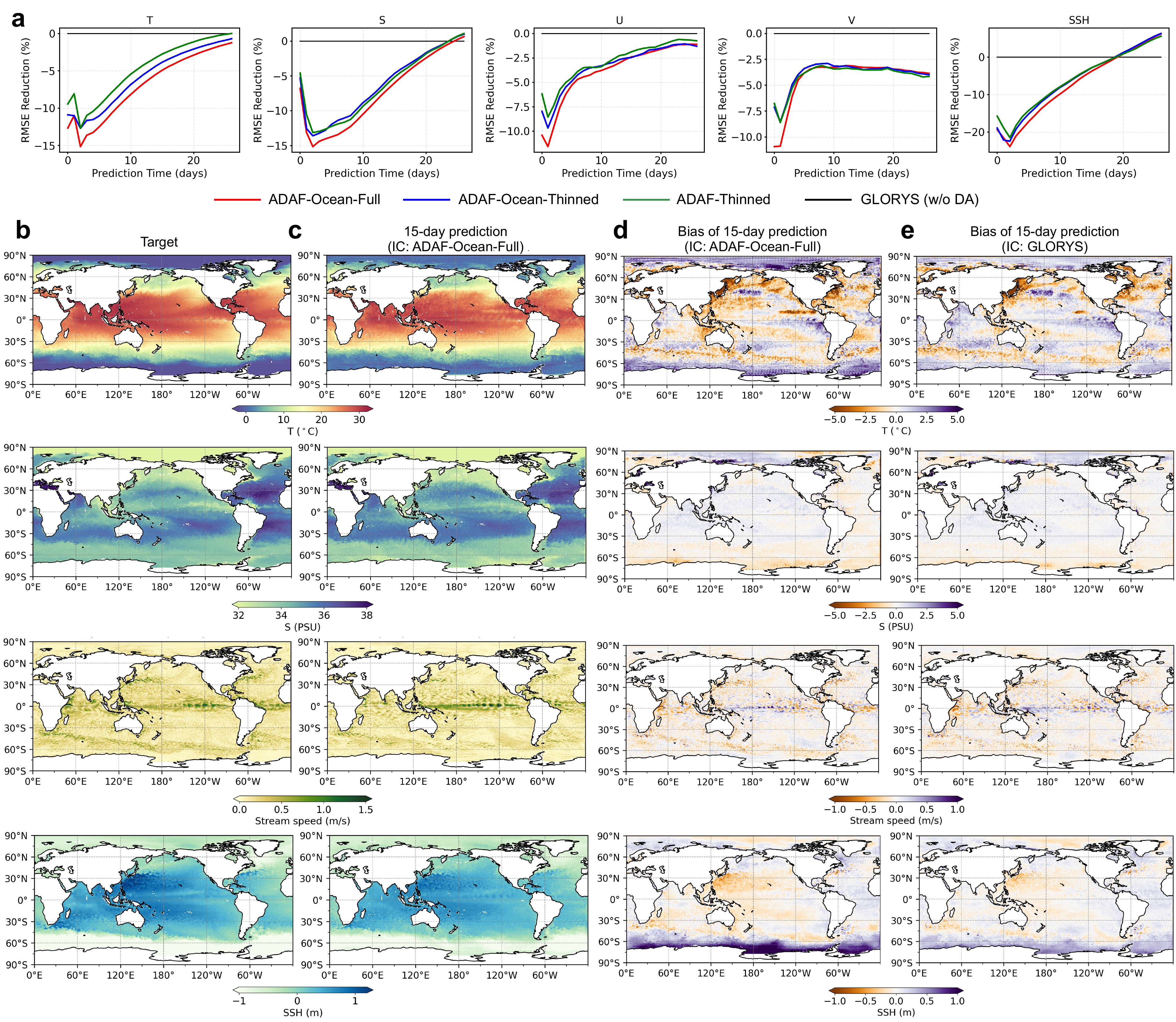}
    \caption{\textbf{Global forecast after a single DA.} \textbf{a} Time-averaged, latitude-weighted RMSE reduction during a 27-day forecast, comparing three types of initial conditions (ICs): \OceanDA-Full (red), \OceanDA-Thinned (blue), and ADAF-Thinned (green), relative to forecasts initialized with baseline GLORYS (w/o DA). More negative values (larger reduction) indicate better forecast skill compared to the baseline. \textbf{b} GLORYS used as a reference target. \textbf{c} 15-day forecast for T, S, Stream speed ($\sqrt{U^2+V^2}$), and SSH using \OceanDA-Full IC. Bias maps of the 15-day forecast using \OceanDA-Full IC (\textbf{d}) and GLORYS IC (\textbf{e}).}
    \label{fig:forecast_skill}
\end{figure}

The primary goal of DA is to provide accurate initial conditions (ICs) to enhance forecast skills. To evaluate the effectiveness of the analysis generated by \OceanDA, we conducted 27-day global forecasts using a DL-based forecast model (details in Supplementary Section~\ref{appsec:forecast_model}). 
We compared forecast skills using three types of ICs: analysis produced by \OceanDA-Full, \OceanDA-Thinned, and ADAF-Thinned. 
The 3–30 days forecasts using GLORYS ICs served as the baseline (w/o~DA), as the 3-day forecast was used as the background in this section.
Fig.~\ref{fig:forecast_skill}{a} shows the time-averaged RMSE reduction, defined as $(\text{RMSE}_{\text{IC}} - \text{RMSE}_\text{w/o~DA}) / \text{RMSE}_\text{w/o~DA}$, over the 27-day forecast for 3 types of IC.

All three DA-initialized configurations demonstrated a clear improvement in forecast skill over the baseline (w/o DA). They achieved substantial initial RMSE reductions (up to -15\% for T and -21\% for SSH) and sustained this skill advantage for at least 20 days across all variables (Fig.~\ref{fig:forecast_skill}{a}). 
Among the DA methods, \OceanDA-Full consistently and significantly outperforms both thinned-data configurations. It exhibits the maximum RMSE reduction at the initial analysis state (t=0), signifying optimal IC improvement.
Importantly, \OceanDA-Thinned also demonstrates a consistent, albeit smaller, advantage over ADAF-Thinned, particularly for T, U, and SSH. This highlights the superior capability of the \OceanDA~framework itself, even when constrained to identical thinned data.
While the skill advantage of all DA runs naturally decays, \OceanDA-Full maintains the most pronounced and sustained superiority for T, S, U, and SSH. Conversely, for V, the skill levels of all three DA methods converge after approximately 10 days.

Furthermore, we analyze a representative 15-day prediction case initialized with \OceanDA-Full ICs (Fig.~\ref{fig:forecast_skill}{c}), which is broadly consistent with the GLORYS reference target (Fig.~\ref{fig:forecast_skill}{b}).
Fig.~\ref{fig:forecast_skill}{d,e} present the bias maps (prediction - target) for 15-day predictions using the two ICs: \OceanDA-Full and GLORYS analysis. For S and stream speed, the spatial distribution of the bias is similar between the two ICs, indicating comparable forecast accuracy.
However, predictions for T and SSH initialized with \OceanDA-Full analysis exhibit larger errors in high-latitude (Arctic and Antarctic) regions than those starting with GLORYS. This is directly attributed to the larger initial errors already present in the \OceanDA-Full ICs (the analysis field) in these polar regions (Fig.~\ref{fig:DA_metrics_with_baselines}{d}).

\section{Discussion}

\OceanDA~represents a paradigm shift in Earth system Data Assimilation (DA) by directly confronting the dual crisis of computational scale and observational data fidelity that has constrained global high-resolution modeling. 
Our framework systematically overcomes the core limitations of both traditional and prior DL methods.
First, \OceanDA~replaces the iterative, high-cost optimization of traditional methods (4DVar/EnKF) with a single, lightweight forward inference pass. This drastically reduces the computational cost. Achieving global 0.25$^\circ$ analysis requires solving a system with only 28 million parameters, a computational scale that is orders of magnitude smaller than the massive linear algebra required by operational NWP-based DA systems. By integrating a purely DL forecasting model, the framework further establishes a computationally efficient prediction system. The analysis fields produced by \OceanDA~provide superior initial conditions, significantly improving forecast skill for up to 20 days.

Second, unlike prior DL-DA methods that fail to address the underlying data fidelity crisis, the Neural Process-inspired architecture in \OceanDA~is inherently designed to handle the irregularity of observations. By transforming raw, multi-source, and multi-scale data into a latent space, we eliminate the necessity for interpolation and thinning. This innovation maximizes the information content of the observational network, as robustly demonstrated by the superior performance of the ADAF-Ocean-Full configuration (Fig.~\ref{fig:DA_metrics_with_baselines}{a,b}) over thinned-data baselines.

\OceanDA~successfully reconstruct 0.25$^\circ$ analysis field from a coarse 1$^\circ$ background, leveraging its AI-driven super-resolution capability. Spectral analysis confirms that the AI-driven reconstruction of the high-wavenumber spectrum demonstrates strong implicit capture of dynamics (Fig.~\ref{fig:high_resolution_DA}{f}). 
This proves the ability of \OceanDA~to physically reconstruct the energy associated with mesoscale and sub-mesoscale features that are lost by traditional linear methods. 
The utility of this high-fidelity reconstruction extends to operational design. 

Our evaluation of observation contributions (Supplementary Section~\ref{appsec:observation_contributions}) provides actionable insights for observation strategies. A nuanced finding emerges for Sea Level Anomaly (SLA) data. Its role is apparently contradictory. The sensitivity analysis (Supplementary Section~\ref{appsec:uncertainty_sensitivity_analysis}) identifies SLA as having the most significant influence, yet the contribution experiment (Supplementary Section~\ref{appsec:observation_contributions}) ranks its unique contribution as less significant.
This discrepancy highlights the different aspects measured, specifically influence versus irreplaceability. The sensitivity test confirms SLA is a primary constraint for geostrophic balance. The contribution test, however, suggests the information from SLA is partially redundant, as the framework implicitly derives sea level components from the density and wind-driven dynamics constrained by SST, SSS, and SSW.

Despite these advancements, several avenues for future research exist to enhance the scientific rigor and utility of \OceanDA.
The current study focuses on surface variables. Extending \OceanDA~to effectively assimilate vertical profiles from in-situ platforms is a critical next step to estimate the subsurface state of the ocean, enhancing its utility for climate and ecosystem modeling.
While the AI-driven reconstruction of the high-wavenumber spectrum demonstrates strong implicit capture of dynamics, our next generation of \OceanDA~will explicitly incorporate soft physical constraints, such as mass, heat, or momentum balance penalties, directly into the loss function to guarantee a physically consistent analysis state.
Moreover, developing a fully coupled framework to simultaneously assimilate oceanic and atmospheric observations, such as wind stress and surface heat flux could significantly improve the representation of complex coupled phenomena, such as tropical cyclones or El Niño events.

\section{Methods}

\subsection{Overview}

The primary goal of \OceanDA~is to reconstruct the state of Ocean by integrating sparse, noisy observational data with background fields derived from forecasting models (Fig.~\ref{fig:model}{a}). This process involves predicting unknown variables at target locations within the global ocean domain $\Omega$, using limited and uncertain information.

Observations are defined as the set of coordinate-value pairs, $\mathbf{O} = \{(\mathbf{x}_i, \mathbf{y}^o_i)\}_{i=1}^{n}$, where $\mathbf{x}_i \in \mathbb{R}^2$ are the spatial coordinates (e.g., latitude and longitude), and $\mathbf{y}^o_i \in \mathbb{R}^{N_O}$ are the observed variables, where $N_O$ is the number of observed variables.
Similarly, the background field is defined as the set $\mathbf{B} = \{(\mathbf{x}_i, \mathbf{y}^b_i)\}_{i=1}^{m}$, where $\mathbf{y}^b_i \in \mathbb{R}^{N_B}$ are the background variables derived from forecasting models, where $N_B$ is the number of background variables.

The objective is to predict the analysis variable $\mathbf{y}^a$ at all target locations $\mathbf{x}^a \in \Omega$ by combining $\mathbf{O}$ and $\mathbf{B}$. Mathematically, this is expressed as:
\begin{equation}
\mathbf{y}^a = F(\mathbf{x}^a \mid \mathbf{O}, \mathbf{B}),
\end{equation}
where $F$ represents the mapping function that fuses sparse, noisy observations with background fields to produce gap-free estimates of the analysis variable $\mathbf{y}^a \in \mathbb{R}^{N_A}$.
Crucially, $F$ is designed to achieve super-resolution, generating higher-resolution analysis fields from coarse-resolution background ($\mathbf{B}$) to resolve fine-scale variability.

\subsection{Model structure}

\begin{figure}[h]
    \centering
    \includegraphics[width=1\linewidth]{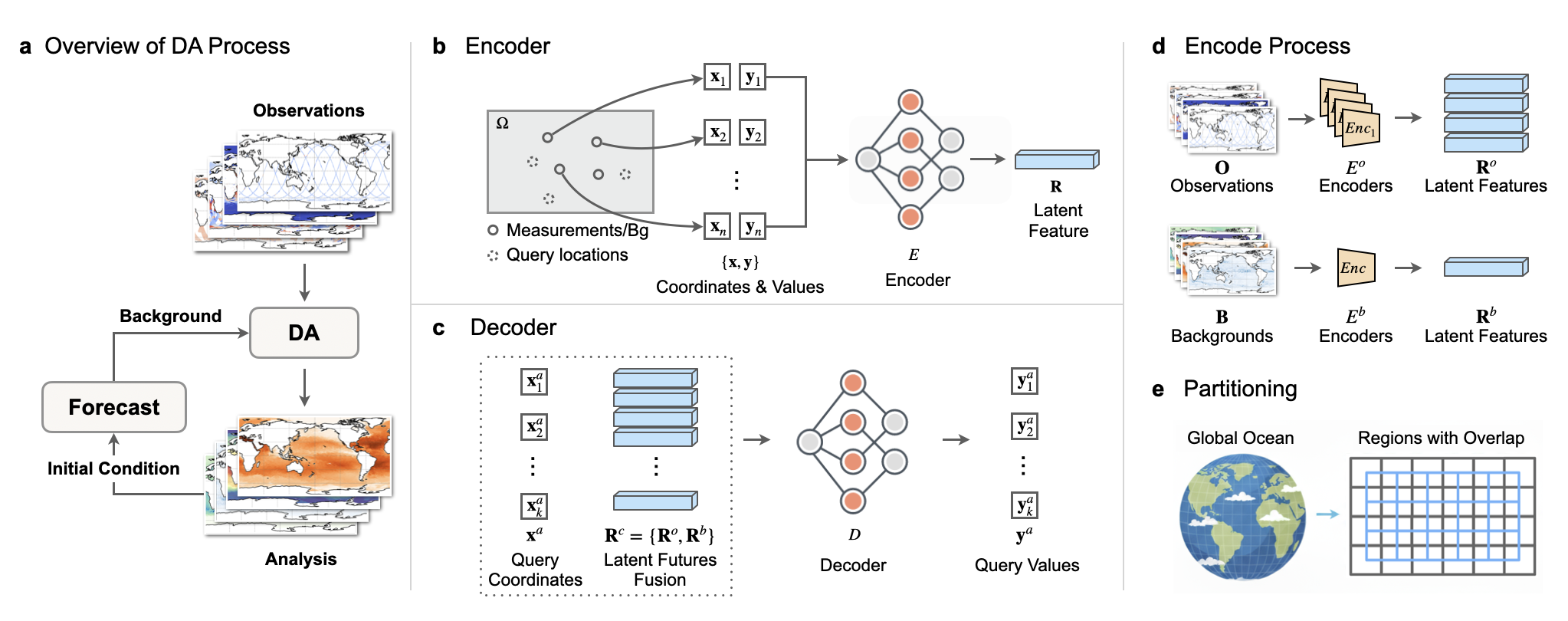}
    \caption{\textbf{The model architecture.} \textbf{a} Overview of the DA process: Observations ($\mathbf{O}$) and background fields ($\mathbf{B}$) are integrated to generate the analysis ($\mathbf{y}^a$). 
    \textbf{b} The point-based Encoder ($E$) processes coordinates and values ($\mathbf{x}, \mathbf{y}$) into the latent feature $\mathbf{R}$.
    \textbf{c} The Decoder ($\mathbf{D}$) uses the fused latent feature ($\mathbf{R}^c$) and query coordinates ($\mathbf{x}^a$) to predict the analysis ($\mathbf{y}^a$). \textbf{d} Specialized encoders ($\{E_j^o\}, E^b$) independently transform multi-source inputs ($\mathbf{O}, \mathbf{B}$) into latent features ($\mathbf{R}^o, \mathbf{R}^b$). \textbf{e} Global partitioning: The domain is split into overlapping regions to ensure continuity and manage computational memory.}
    \label{fig:model}
\end{figure}

\OceanDA~employs an encoder-decoder architecture inspired by Neural Processes (NPs)\cite{garnelo2018}, which are particularly well-suited for handling irregular, sparse, and multi-scale environmental sensor data\cite{Andersson2023, Allen2025, Vaughan2022}.
To implement the NP-inspired framework, we utilize Multi-Layer Perceptrons (MLPs)~\cite{Hornik1089} as the core building blocks for their simplicity, computational efficiency, and adaptability. MLPs encode spatial coordinates and variable values into latent representations, enabling seamless integration of multi-source data, such as satellite swaths and in-situ measurements, while preserving the fidelity of the original observations.

The model takes three inputs: observations ($\mathbf{O}$), background fields ($\mathbf{B}$), and target analysis coordinates ($\mathbf{x}^a$). It learns a continuous, non-linear mapping $F$ to estimate the analysis variables ($\mathbf{y}^a$) at any desired grid location:
\begin{equation} 
    \mathbf{y}^a = F(\mathbf{x}^a | \mathbf{O}, \mathbf{B}) 
\end{equation}
 
To implement this mapping, the architecture features a set of specialized encoders, $\left\{E_j^o\right\}$ for observations and $E^b$ for the background field, to handle heterogeneous data formats independently (Fig.~\ref{fig:model}{d}).
Crucially, while all encoders ($\left\{E_j^o\right\}$ and $E^b$) share a conceptually similar MLP-based architecture, including (e.g., depth, activation functions, and skip connections), their input layers are specifically tailored. The dimensionality of the first layer of each encoder is adapted to match the specific number of variables in its corresponding data source, allowing them to consistently process heterogeneous input data organized as isolated points.

To achieve this structural homogeneity, both the sparse observations ($\mathbf{O}$) and the gridded background field ($\mathbf{B}$) are transformed into a combined point vector $(\mathbf{x}, \mathbf{y})$ before encoding.
For the sparse observations ($\mathbf{O}_j$), the measurements ($\mathbf{y}$) are first combined with its trigonometrically encoded coordinates $\mathbf{x}$ (refer to Section~\ref{sec:partition_location_encoding}, Fig.~\ref{fig:model}{b}). For the gridded background field ($\mathbf{B}$), the field is first sampled or flattened into discrete data points $\mathbf{y}$ at their respective grid locations $\mathbf{x}$.

This combined vector $(\mathbf{x}, \mathbf{y})$ is then transformed by the corresponding encoder:
\begin{equation}
\mathbf{R} = E({\mathbf{x}, \mathbf{y}}),
\end{equation}
where $\mathbf{R}$ is the latent feature that represents the contextual information of the input data, and $E$ denotes the general MLP-based encoding function.
The latent features for observations ($\mathbf{R}_j^o$) and background fields ($\mathbf{R}^b$) are computed separately using their dedicated encoders:
\begin{equation}
\mathbf{R}_j^o = E_j^o(\mathbf{O}_j), \quad \mathbf{R}^b = E^b(\mathbf{B}),
\end{equation}
Observations ($\mathbf{O}_j$) consist of a variable number of sparse measurements and are processed by separate encoders ($E_j^o$), producing the latent feature ($\mathbf{R}_j^o$). $j$ indexes different observation sources. These encoders are crucial for direct DA as they naturally handle the irregular, sparse input structure.
The dedicated encoder, $E^b$, processes the background field and encodes its spatial and physical context, producing the latent feature $\mathbf{R}^b$.

The decoder ($\mathbf{D}$, Fig.~\ref{fig:model}{c}) performs the final state estimation and drives resolution enhancement.
The fused latent feature, $\mathbf{R}^c = \{ \mathbf{R}^b, \mathbf{R}^o\}$, combines all encoded contextual information into a single representation, which is then fed into the decoder.
The decoder takes three inputs: the target query coordinates ($\mathbf{x}^a$), and the separate latent features for observations ($\mathbf{R}^o$) and background fields ($\mathbf{R}^b$). The analysis variable ($\mathbf{y}^a$) at the queried locations is computed as:
\begin{equation}
\mathbf{y}^a = D(\mathbf{x}^a, \mathbf{R}^o, \mathbf{R}^b)
\end{equation}
The final output, $\mathbf{y}^a$, represents the analysis variable at the precisely defined latitude and longitude coordinates ($\mathbf{x}^a$).
By querying coordinates on a high-resolution grid, the architecture inherently achieves its super-resolution capability. 

Both the encoders ($\{E_j^o\}, E^b$) and the decoder ($\mathbf{D}$) are implemented as deep MLPs, featuring linear layers, activation functions, and skip connections for enhanced gradient flow and feature mixing. 
This design—processing feature vectors rather than image grids—enables the direct assimilation of irregular and multi-source data without requiring prior interpolation or resampling.

\subsection{Partition and location encoding}\label{sec:partition_location_encoding}

To efficiently process global-scale, high-dimensional data, \OceanDA~partitions the global ocean into overlapping sub-regions (patches) (Fig.~\ref{fig:model}{e}). This overlapping design is critical for ensuring that boundary areas are fully captured, thereby preventing discontinuities between regions.
This patch-based strategy is crucial for the scalability of \OceanDA, primarily by decoupling the instantaneous GPU memory footprint from the total global domain size. Without partitioning, processing the entire global field would require loading all input tensors simultaneously, which is computationally infeasible.
By splitting the domain patch-by-patch, the GPU memory requirement is determined only by the size of a single patch, not the global field. This allows the model to scale to arbitrarily high resolutions  without increasing the instantaneous GPU memory demand. 
A further critical benefit of partitions is implicit data augmentation. 
A single global state is split into numerous (overlapping) patches, each serving as an independent training sample. This massively inflates the training dataset, enabling the model to learn from a relatively short temporal record.

The model does not inherently account for spatial locations. To incorporate geospatial information, we encode the latitude and longitude into a continuous, cyclic, and normalized representation using trigonometric transformations. This ensures the preservation of the cyclic properties of geospatial coordinates (e.g., the periodicity of longitude) and improves the numerical stability of training. The latitude $\phi_{\text{rad}}$ and longitude $\lambda_{\text{rad}}$, expressed in radians, are transformed as follows:
\begin{align}
    \text{lat\_sin} &= \sin(\phi_{\text{rad}}), \quad \text{lat\_cos} = \cos(\phi_{\text{rad}}), \\
    \text{lon\_sin} &= \sin(\lambda_{\text{rad}}), \quad \text{lon\_cos} = \cos(\lambda_{\text{rad}}).
\end{align}
These transformations provide a cyclically consistent encoding of spatial locations, enabling the model to effectively represent the continuity of geospatial relationships. By combining the partitioning strategy with trigonometric location encoding, the model can construct field predictions at arbitrary subsets of the global domain. This flexibility significantly reduces computational requirements, as predictions can be generated piece by piece rather than processing the entire domain at once.

\subsection{Datasets}\label{sec:datasets}

To demonstrate the efficacy of ADAF-Ocean in global surface ocean DA tasks, we utilized a combination of a high-resolution reanalysis product for ground truth and six diverse real-world observational datasets.
We use the GLORYS12V1~\cite{glorys, glorys2021} dataset (referred to as GLORYS), a global ocean physics reanalysis product from the Copernicus Marine Environment Monitoring Service (CMEMS). GLORYS, based on the Nucleus for European Ocean Models (NEMO), assimilates various observations to reconstruct the 3D ocean circulation. The native spatial resolution of the product is $1/12^\circ$.
In this study, we focus on five key surface variables: temperature (T), salinity (S), zonal velocity (U), meridional velocity (V), and sea surface height (SSH). These variables were temporally aggregated to a daily resolution.

The study assimilates six types of multi-source observations, classified into satellite-based and in-situ measurements (Table~\ref{tab:observations}). Visualization examples are presented in Supplementary Fig.~\ref{appfig:obs_vis}.
We utilize the high-resolution, near-real-time Sea Surface Temperature (SST) from the Advanced Very High Resolution Radiometer (AVHRR) (4 km)~\cite{Saha2016} to constrain thermal dynamics. Sea Level Anomaly (SLA) from Jason-2/3 (25 km)~\cite{CMEMS_SSH} provides crucial information for geostrophic balance and circulation features. For air-sea interaction, Sea Surface Wind (SSW) from MetOp-A (25 km)~\cite{CMEMS_SSW} constrains surface currents. Finally, Sea Surface Salinity (SSS) from SMOS (25 km)~\cite{Nicolas2016} and Sea Ice Concentration (SIC) from NOAA/NSIDC (100 km)~\cite{Meier2021} are included to constrain density-driven processes and polar dynamics, respectively. This combination of multi-resolution satellite inputs is directly leveraged by the model's architecture to maximize information utilization. 
We also employ the HadIOD.1.2.0.0 dataset~\cite{Atkinson2014}, which includes sparse, point-based measurements of surface and subsurface temperature (T) and salinity (S), sourced from ships, buoys, and profiling floats. This dataset is quality-controlled and contains metadata concerning bias corrections and uncertainties. The integration of high-resolution, globally-covered satellite data with local in-situ measurements provides a robust foundation for accurate ocean state estimation.

\begin{table}[h!]
\caption{\textbf{Summary of Observational Datasets}}\label{tab:observations}
\centering
\begin{tabular*}{\textwidth}{@{\extracolsep\fill}cccc}
\toprule
\textbf{Observation} & \textbf{Source} & \textbf{\makecell[c]{Spatial\\Resolution}} & \textbf{\makecell[c]{Temporal\\Resolution}} \\ 
\midrule
SST(SAT) & AVHRR~\cite{Saha2016} & 4 km & Twice-daily  \\ 
SSS(SAT) & SMOS~\cite{Nicolas2016} & 25 km & Twice-daily \\ 
SSW(SAT) & MetOp-A~\cite{CMEMS_SSW} & 25 km & Twice-daily \\ 
SIC(SAT) & NOAA/NSIDC~\cite{Meier2021} & 100 km & Daily \\ 
SLA(SAT) & Jason-2,3~\cite{CMEMS_SSH} & 25 km & Daily  \\ 
T, S(In-situ) & HadIOD.1.2.0.0~\cite{Atkinson2014} & Point-based & Daily \\ 
\botrule
\end{tabular*}
\footnotetext{‘SAT’ refers to satellite observations, while ‘In-situ’ refers to measurements collected from ships, buoys, and floats. All satellite observations are Level-3 (L3) products.}
\end{table}


\subsection{Model training}\label{sec:model_training}

To establish a purely AI-based DA framework, we replace the traditional numerical weather prediction (NWP) model with an AI forecast model. The AI forecast serves as the background input for the \OceanDA. In this study, we employ Triton~\cite{Wu2025Triton} as the forecast model, a hierarchical architecture designed to process information across multiple spatial and temporal resolutions. Triton mitigates spectral bias and explicitly captures cross-scale dynamics, enabling it to achieve superior performance in various challenging forecasting tasks. For training the forecast model, we use the GLORYS reanalysis dataset, covering the period from 2010 to 2020. Training is conducted on data from 2010 to 2016, with 2017 data reserved for validation. Forecasts are performed for the independent period from 2018 to 2020, using a forecast step of one day. During inference, Triton operates in an auto-regressive manner, where initial conditions are sourced either from reanalysis fields or from the analysis produced by \OceanDA.

The training data for the \OceanDA~model consists of input-target pairs. The inputs include background fields and observations, while the target is the analysis increment (the bias between the GLORYS reanalysis and the background fields). This setup allows the model to focus on learning the necessary corrections to improve the background. The assimilation time window spans one day, including data from the day prior to the analysis time. Background fields are sourced from the third day predictions of the AI forecast model. To avoid temporal data leakage, the dataset is split chronologically, with 2018 data used for training, 2019 for validation, and 2020 for testing.

The spatial domain of the target and background fields is $180^\circ \times 360^\circ$ with a spatial resolution of 1$^\circ$. These fields are partitioned into patches of size $10^\circ \times 20^\circ$ with overlapping $5^\circ \times 5^\circ$. 
Observation patches correspond to the same regions as the target patches, but their sizes vary depending on their resolution. Specifically, the patch sizes are (250, 500) for SST, (50, 100) for SSS, (40, 80) for SSW, and (10, 20) for both SLA and SIC. In-situ observations are also partitioned into patches of $10^\circ \times 20^\circ$. Each variable has a specific number of channels: SST, SSS, SLA, and SIC each have one channel, while SSW includes two channels for wind direction and wind speed. 

The \OceanDA~model is implemented using the PyTorch\cite{pytorch} framework. The encoder and decoder architectures of the model are configured with depths of 6 and 8 layers, respectively. 
Training is conducted on two NVIDIA A100 GPUs and requires approximately 24 hours for 200 epochs. A masked-land Mean Squared Error (MSE) loss function is employed, which excludes land areas to ensure the model focuses exclusively on oceanic regions. The optimizer used is Adam\cite{Adamw}, and a MultiStepLR scheduler is applied to decay the learning rate by a factor of 0.5 at epochs 80, 160, and 200. 

\subsection{Evaluation metrics}

To evaluate the performance of the proposed model, we used three key metrics commonly applied in geophysical data analysis: latitude-weighted Root Mean Squared Error (RMSE), Mean Absolute Error (MAE), and latitude-weighted Anomaly Correlation Coefficient (ACC)~\cite{wang2024accurate, Xu2025, xiong2023ai}.

Latitude-Weighted RMSE quantifies the difference between predicted and true values while accounting for the uneven spatial distribution of grid points across latitudes, which is calculated as follows:
\begin{equation}
    \text{RMSE} = \sqrt{\frac{\sum_{i=1}^{H} \sum_{j=1}^{W} \cos{\phi_i} (\hat{y}_{i,j} - y_{i,j})^2}{\sum_{i=1}^{H} \sum_{j=1}^{W} \cos{\phi_i}}}
    \label{eq:rmse}
\end{equation}
where $H$ and $W$ represent the total number of latitude and longitude grid points, $\hat{y}$ and $y$ are the analysis produced by \OceanDA~and the reanalysis references from GLORYS, respectively, and $\cos{\phi_i}$ is the weight applied at latitude $\phi_i$.

MAE is primarily used to quantify the spatial distribution of errors at the grid level. It is less sensitive to large outliers than RMSE and is calculated as the unweighted average of absolute errors over $N$ grid points:
\begin{equation}
    \text{MAE} = \frac{1}{H \times W} \sum_{i=1}^{H} \sum_{j=1}^{W} |\hat{y}_{i,j} - y_{i,j}| 
    \label{eq:mae}
\end{equation}
where $N$ is the total number of valid grid points in the region of interest.

ACC measures the correlation between predicted and true anomalies, highlighting the ability to capture spatial deviations from the mean state. Anomalies are calculated by subtracting the climatological mean. The latitude-weighted ACC is formulated as:
\begin{equation}
    \text{ACC} = \frac{\sum_{i=1}^{H} \sum_{j=1}^{W} \cos{\phi_i} (y_{i,j} - c_{i,j})(\hat{y}_{i,j} - c_{i,j})}{\sqrt{\left[ \sum_{i=1}^{H} \sum_{j=1}^{W} \cos{\phi_i} (y_{i,j} - c_{i,j})^2 \right] \left[ \sum_{i=1}^{H} \sum_{j=1}^{W} \cos{\phi_i} (\hat{y}_{i,j} - c_{i,j})^2 \right] }}
\label{eq:acc}
\end{equation}
where $c$ represents the climatological mean over the day-of-year, which is averaged from 2010 to 2020 with GLORYS data. $\cos{\phi_i}$ is the latitude weight.

\section{Data Availability}

The datasets used in this study are publicly available from the following sources: (i) The GLORYS reanalysis dataset is available from the Copernicus Marine Service at https://doi.org/10.48670/moi-00021.
(ii) The AVHRR Pathfinder SST is available from NCEI at https://doi.org/10.7289/v52j68xx.
(iii) The SMOS L3 Sea Surface Salinity (MULTIOBS\_GLO\_PHY\_SSS\_L3\_MYNRT\_015\_014) is available from the Copernicus Marine Service at https://doi.org/10.1016/j.rse.2016.02.061.
(iv) The sea surface wind observations from ASCAT (WIND\_GLO\_PHY\_L3\_MY\_012\_005) is available at https://doi.org/10.48670/moi-00183.
(v) The sea ice concentration Climate Data Record (CDR) is available from NSIDC at https://doi.org/10.7265/efmz-2t65.
(vi) The global ocean along track L3 sea surface heights (SEALEVEL\_GLO\_PHY\_L3\_MY\_008\_062) can be accessed at https://doi.org/10.48670/moi-00146.
(vii) The HadIOD version 1.2.0.0
can be accessed at https://www.metoffice.gov.uk/hadobs/hadiod.
Some example datasets and the weights of the trained model can be accessed from~\cite{xiang_2025}.

\section{Code Availability}

The core code of ADAF-Ocean is available via GitHub at https://github.com/xiangyanfei212/ADAF.git.

\clearpage
\bibliography{sn-bibliography}

\section{Acknowledgements}

We sincerely appreciate the researchers and staff at the Met Office (for the HadIOD dataset), GHRSST and the NOAA National Centers for Environmental Information (for SST), the Copernicus Marine Environment Monitoring Service (for SSS, SSW, SLA, and the GLORYS reanalysis), and the NOAA/NSIDC (for the SIC Climate Data Record). Their significant efforts in curating, maintaining, and providing these public datasets were essential for our experiments and analyses.


The authors are grateful for the support of Natural Science Foundation of China (No: 12427811(X.L.), 52176082(H.H.), 12275261(Q.L.)), and the authors also want to appreciate the LingChuang Research Project of China National Nuclear Corporation (H.H.) and “Shu Guang” project supported by Shanghai Municipal Education Commission (X.L.) and Shanghai Education Development Foundation (21SG13(H.H.)). In addition, the authors are especially grateful to Professor Jinbiao Xiong for his guidance and advice.

\section{Contributions}

S.W., J.C. and H.H. conceived the study. S.W., J.C., X.L., T.Z., X.C., Q.L., D.S. and H.H. designed the study. S.W. and J.C. developed the program. S.W., J.C., X.L. and H.H. performed the analyses. X.L., T.Z., X.C., Q.L. and D.S. supervised the analyses. S.W. and H.H. wrote the paper with input from all authors.

\clearpage
\begin{appendices}

\section{Performance of analysis}

\begin{figure}[h]
    \centering
    \includegraphics[width=1\linewidth]{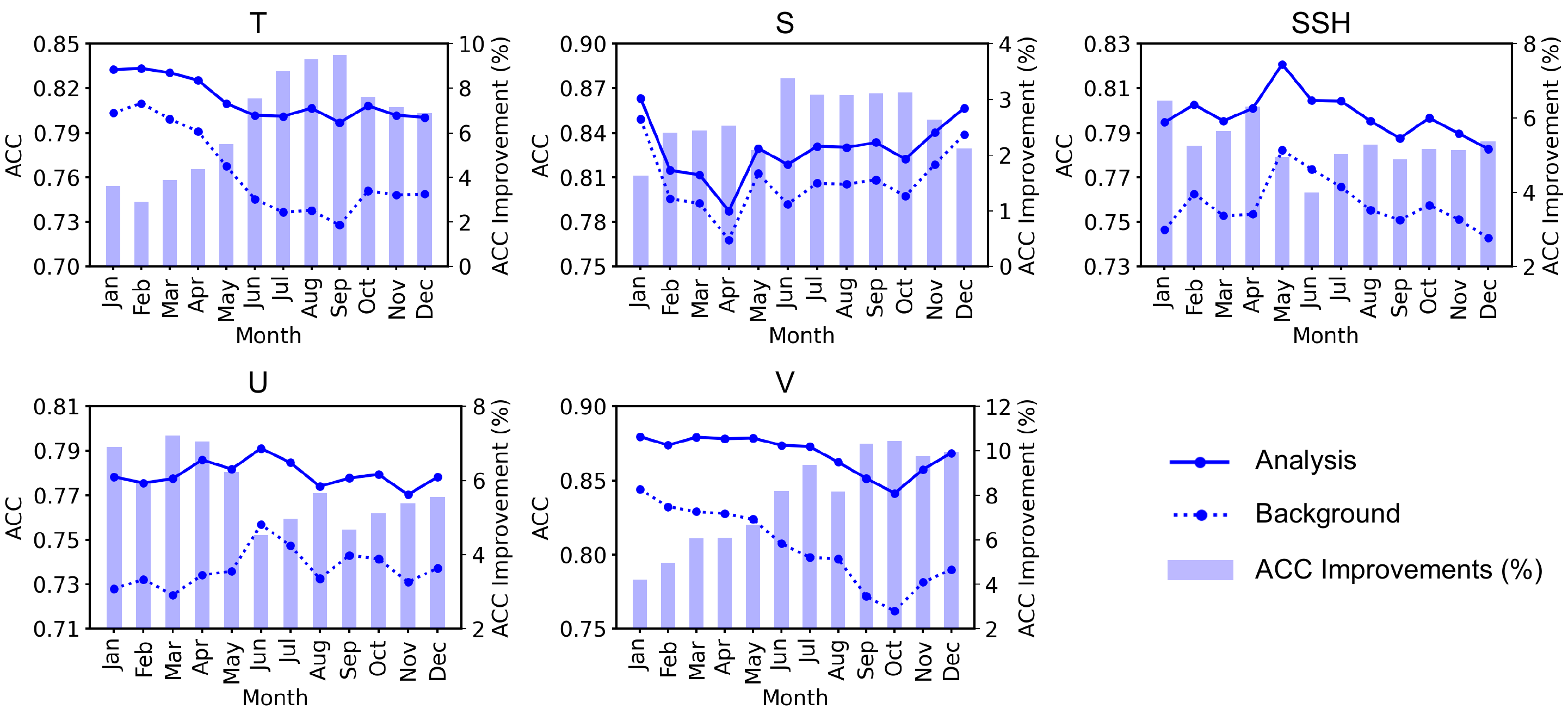}
    \caption{\textbf{Monthly ACC and Improvements for 5 Ocean Surface Variables.} Solid lines represent the analysis (\OceanDA), dashed lines indicate the background (no DA), and bars show ACC improvements for temperature (T), salinity (S), zonal velocity (U), meridional velocity (V), and sea surface height (SSH).}
    \label{appfig:monthly_acc}
\end{figure}

\section{Observations contributions}\label{appsec:observation_contributions}

\begin{figure}[h]
   \centering
   \includegraphics[width=1\linewidth]{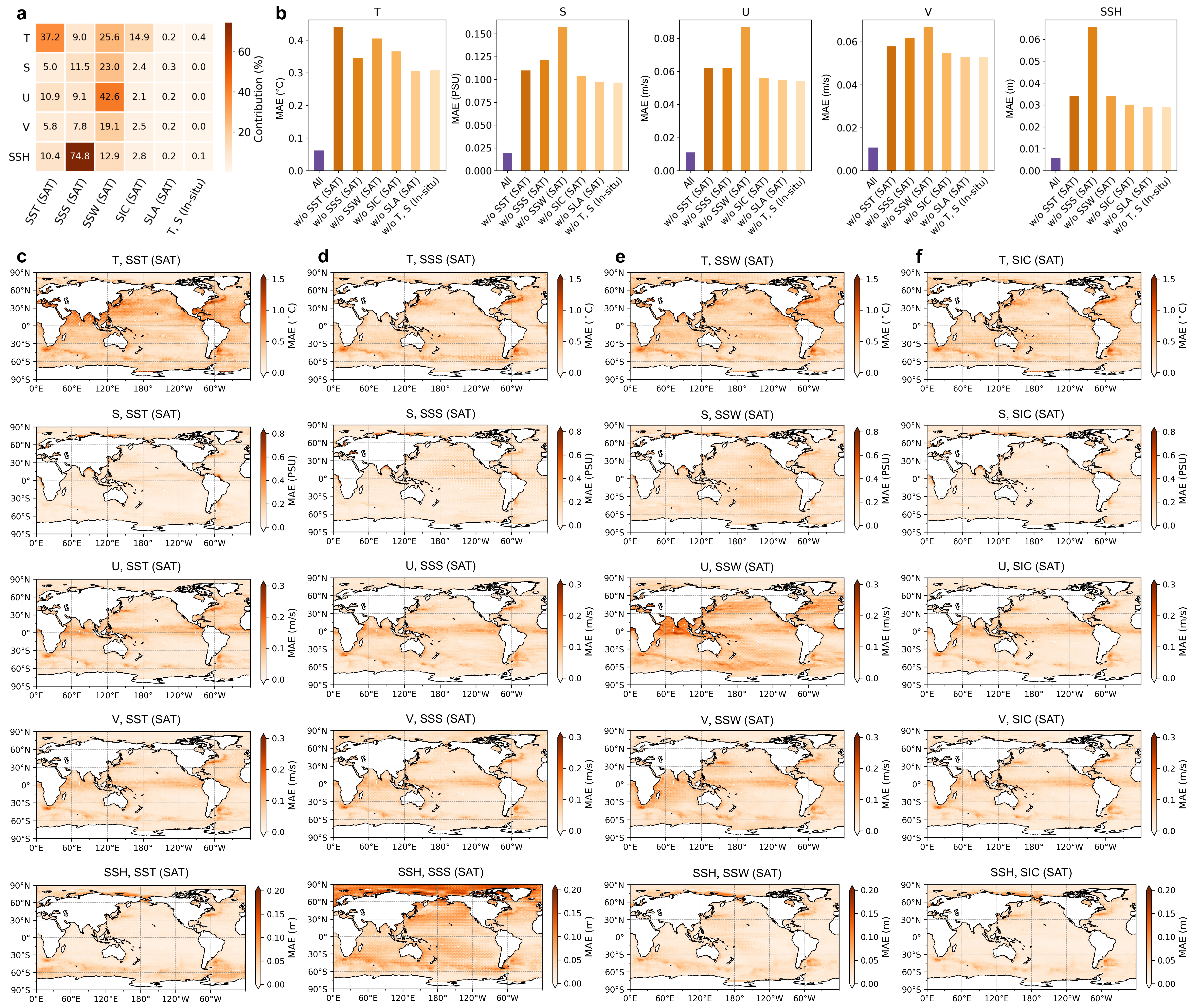}
   \caption{\textbf{Observation Contribution.} \textbf{a} Relative contributions of 5 observational datasets: satellite-measured (SAT) Sea Surface Temperature (SST), Salinity (SSS), Wind (SSW), Ice Concentration (SIC), and Sea Level Anomaly (SLA), along with in-situ temperature (T) and salinity (S) measurements. Contribution is quantified as $(MAE_{\text{w/o}~\text{obs}} - MAE_{\text{all}}) / MAE_{\text{all}}$, where higher values indicate greater importance. \textbf{b} Global average MAE when specific observational datasets are excluded (w/o). \textbf{c-f} Spatial MAE reductions achieved by assimilating satellite-based observations.}
   \label{appfig:obs_contrib}
\end{figure}

Observations are critical for DA, as they provide the direct measurements of oceanic conditions. 
To quantify the contributions of different observation types, we systematically evaluate the impact of excluding specific observations and measure their influence on reducing mean absolute error (MAE) (Fig.~\ref{appfig:obs_contrib}{b}). The "All" scenario represents the baseline, where all types of observations are assimilated. "w/o SST (SAT)" indicates the exclusion of satellite-based SST observations during DA, and similar labels apply to other observation types. The contributions of each observation type across five variables are quantified in Fig.~\ref{appfig:obs_contrib}{a}.

Satellite-based SST observations make the largest contribution to improving the estimation of T, particularly in western boundary currents such as the Kuroshio Current and the Gulf Stream. For S, satellite-based SSW observations have the greatest impact, especially in the Western Pacific (Fig.~\ref{appfig:obs_contrib}{e}). This is because SSW drives critical physical processes, such as ocean circulation, evaporation and precipitation, surface mixing, and the redistribution of freshwater fluxes, all of which influence SSS. Although satellite-based SSS observations provide direct salinity measurements, their impact is more localized and limited by spatial and temporal coverage. In contrast, SSW observations indirectly enhance SSS estimation by constraining the physical processes that govern salinity distribution on a global scale, making them more essential for surface salinity estimation.

Ocean surface currents (U and V) also benefit most from satellite-based SSW observations, with notable improvements observed in the North Indian Ocean, North Pacific Ocean, and Southern Ocean (Fig.~\ref{appfig:obs_contrib}{e}). This reflects the critical role of winds in driving surface currents and shaping global circulation patterns. For SSH, satellite-based SSS observations are the most influential (Fig.~\ref{appfig:obs_contrib}{d}). Improvements are evident across nearly all ocean regions, with particularly pronounced enhancements in the Arctic Ocean. This underscores the importance of salinity in affecting ocean density and sea level variability. In contrast, satellite-based SLA observations are less significant, as they primarily capture short-term, localized sea level changes rather than density-driven processes that govern SSH.

The results highlights the vital role of satellite-based observations—particularly SST, SSW, and SSS—in improving the accuracy of ocean state estimation. By quantifying their contributions, we provide actionable insights for optimizing observation strategies, especially in dynamic and climatically significant regions such as the Arctic Ocean, the Western Pacific, and major current systems. These findings pave the way for advancing DA frameworks and enhancing global ocean estimation capabilities.

\section{Uncertainty and sensitivity}\label{appsec:uncertainty_sensitivity_analysis}

Accurately accounting for errors and uncertainties in observations is essential for effective DA and reliable state estimation. To consider these uncertainties and evaluate their impact on the analysis, we conducted a series of perturbation experiments.
Each type of satellite observational dataset—sea surface temperature (SST), sea surface salinity (SSS), wind speed, and sea level anomaly (SLA)—was perturbed 50 times by adding random Gaussian noise. For each perturbation, the standard deviation of the noise was sampled from the climatological standard deviation of the respective observation type, as described mathematically:
\begin{equation}
O_{\text{perturbed},i} = O_{\text{original}} + \mathcal{N}(0, \sigma_i^2),
\label{eq:perturbed_observation}
\end{equation}
where $\sigma_i$ represents the observational error sampled from the climatological standard deviation, and $i$ indicates the perturbation instance. 
The sensitivity of each variable at each grid point was quantified as the standard deviation of the 50 analysis fields:
\begin{equation}
\text{Sensitivity}(lat, lon) = \operatorname{std} \left( { V_i(lat, lon) }_{i=1}^{50} \right),
\label{eq:sensitivity}
\end{equation}
where $V_i(lat, lon)$ represents the value of variable $V$ at grid location $(lat, lon)$ in the $i$-th analysis field. To make the sensitivity metrics dimensionless and comparable across different variables, the computed sensitivities were normalized by dividing them by the climatological standard deviation of the corresponding observation type.

\begin{figure}[h]
    \centering
    \includegraphics[width=1\linewidth]{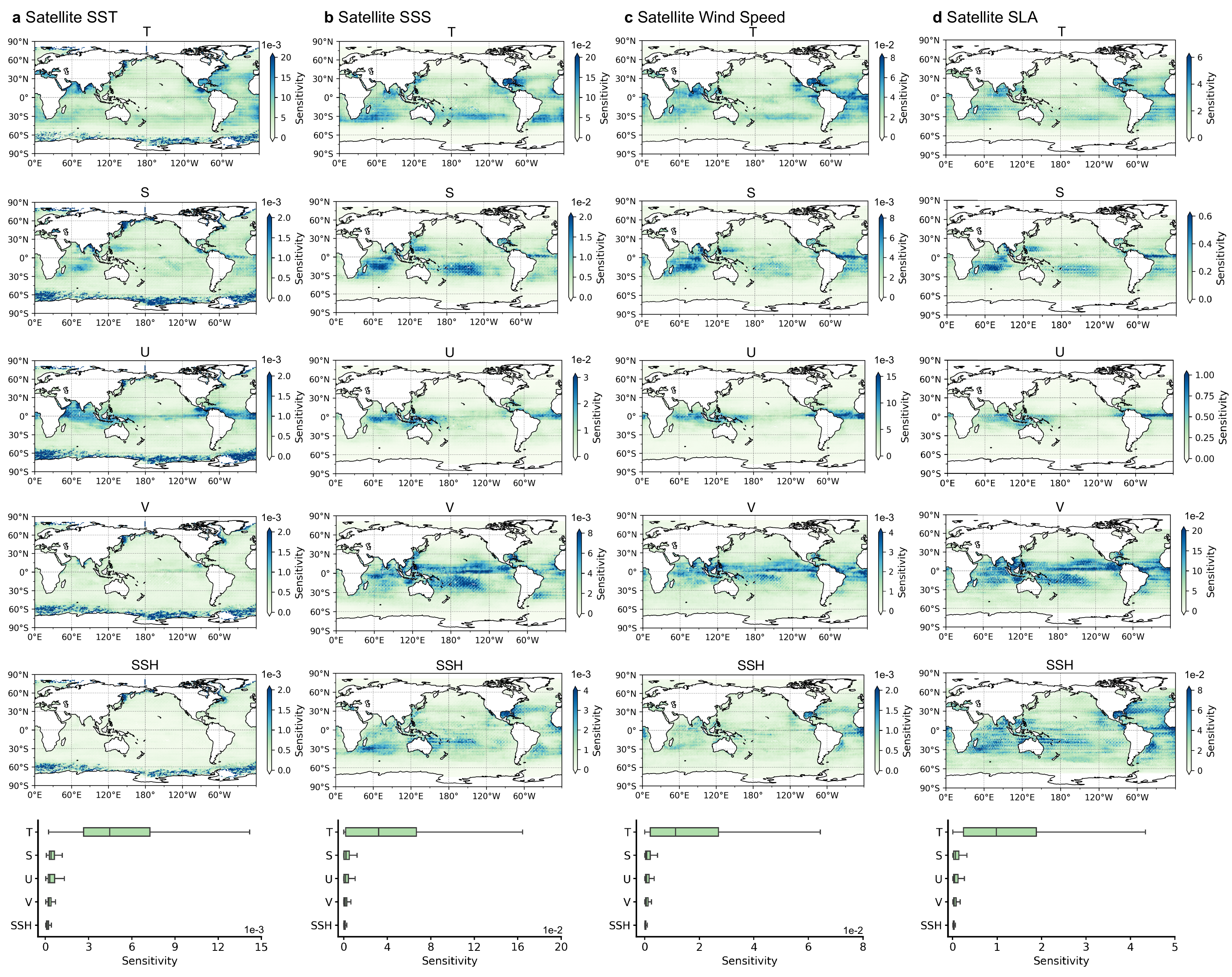}
    \caption{\textbf{Sensitivity to Satellite Observations.} Columns a-d present the sensitivities of the analysis fields to satellite observations of SST, SSS, wind speed, and SLA. Rows correspond to different variables (T, S, U, V, SSH), with the last row summarizing the overall sensitivity distributions across all variables. 
    }
    \label{appfig:unc_sensitivity}
\end{figure}

Fig.~\ref{appfig:unc_sensitivity} illustrates the sensitivity of state estimation to four types of satellite observations. The SST analysis field (T) exhibits notable sensitivity to all observation types, which can be attributed to its strong coupling with SST and related processes, such as air-sea heat flux and ocean mixing. Sensitivity is particularly high near land-sea boundaries, reflecting the complexity and variability of physical processes in these regions. 
In contrast, the SSH analysis field shows the lowest sensitivity. This reduced sensitivity can be attributed to the strong dynamical constraints imposed by geostrophic balance and the assimilation system. These constraints make SSH less dependent on additional observations compared to T, U, and V, indicating that it is already well-constrained in the current system.

Among the four observation types, SLA has the most significant influence on the analysis fields (see Fig.~\ref{appfig:unc_sensitivity}{d}). Perturbations to SLA have a widespread impact on SSH, affecting both regional and basin-wide scales. This highlights the critical role of satellite SLA in providing information for geostrophic balance and large-scale ocean current adjustments. 
The sensitivity maps for the zonal (U) and meridional (V) currents velocity fields reveal strikingly similar spatial patterns across three satellite observation types—SSS, wind speed, and SLA. High-sensitivity regions are concentrated along the equator, underscoring the importance of satellite observations in constraining equatorial currents. These currents, such as the Equatorial Undercurrent and zonal jets, are vital for global ocean circulation and coupled ocean-atmosphere dynamics. 

These results underscore the high sensitivity of  satellite observations, particularly SLA, in ocean state estimation. By identifying key regions and variables with high sensitivity, this perturbation analysis provides a foundation for optimizing observational strategies and enhancing global ocean monitoring systems.

\section{High resolution data assimilation (HR-DA)}\label{sec:high_res_da}

\begin{figure}
    \centering
    \includegraphics[width=1\linewidth]{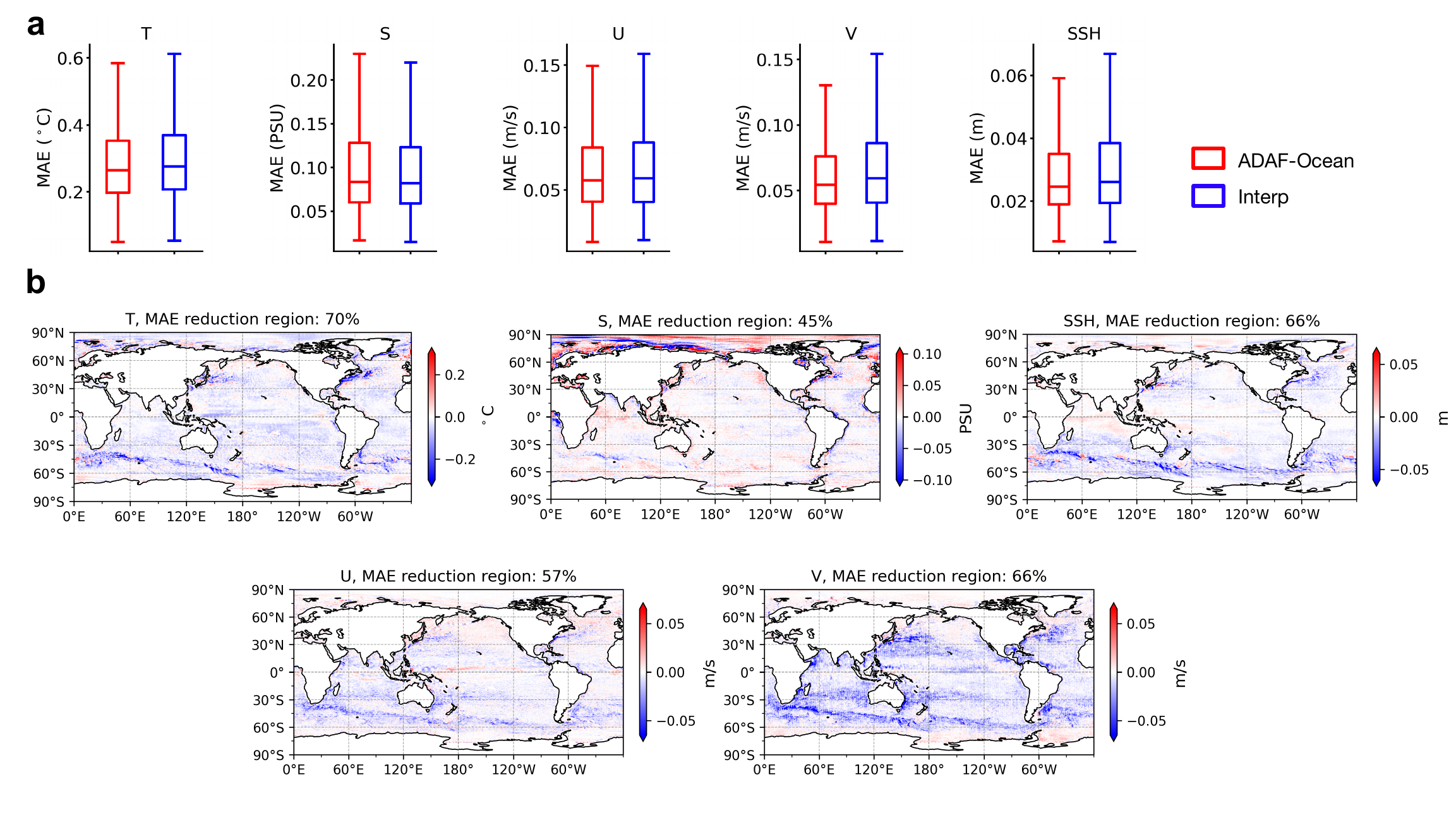}
    \caption{\textbf{Global Performance Assessment of High-Resolution Data Assimilation (HR-DA).} \textbf{a} Global distribution of Mean Absolute Error (MAE) for five surface variables (T, S, U, V, SSH) comparing \OceanDA~and linear interpolation (Interp). \textbf{b} Spatial distribution of MAE reduction ($\text{MAE}_{\text{\OceanDA}} - \text{MAE}_{\text{Interp}}$). Negative values (blue) indicate MAE reduction achieved by \OceanDA. The percentage of the global grid points showing MAE reduction is 70\% for T, 45\% for S, 66\% for SSH, 57\% for U, and 66\% for V.}
\label{appfig:high_resolution_DA_mae}
\end{figure}

Satellite observation resolution plays a critical role in high-resolution data assimilation (HR-DA). To evaluate its importance, we conducted experiments using observations at three resolutions: the original resolution (0.04$^\circ$ for SST, 0.2$^\circ$ for SSS, and 0.25$^\circ$ for SSW), 0.5$^\circ$, and 1$^\circ$. For fair comparisons, separate models were trained for each resolution to ensure unbiased results.
To quantify the impact of resolution, we calculated the latitude-weighted RMSE reduction ratio, which measures the relative improvement in RMSE when switching from lower- to original-resolution observations:
\begin{equation}
    \text{RMSE Reduction} = \frac{\text{RMSE}_{lr} - \text{RMSE}_{origin}}{\text{RMSE}_{lr}}
    \label{eq:diff_res_impact}
\end{equation}
Here, $\text{RMSE}{lr}$ and $\text{RMSE}_{origin}$ represent the RMSE for lower- and original-resolution observations, respectively.

\begin{figure}[h]
    \centering
    \includegraphics[width=1\linewidth]{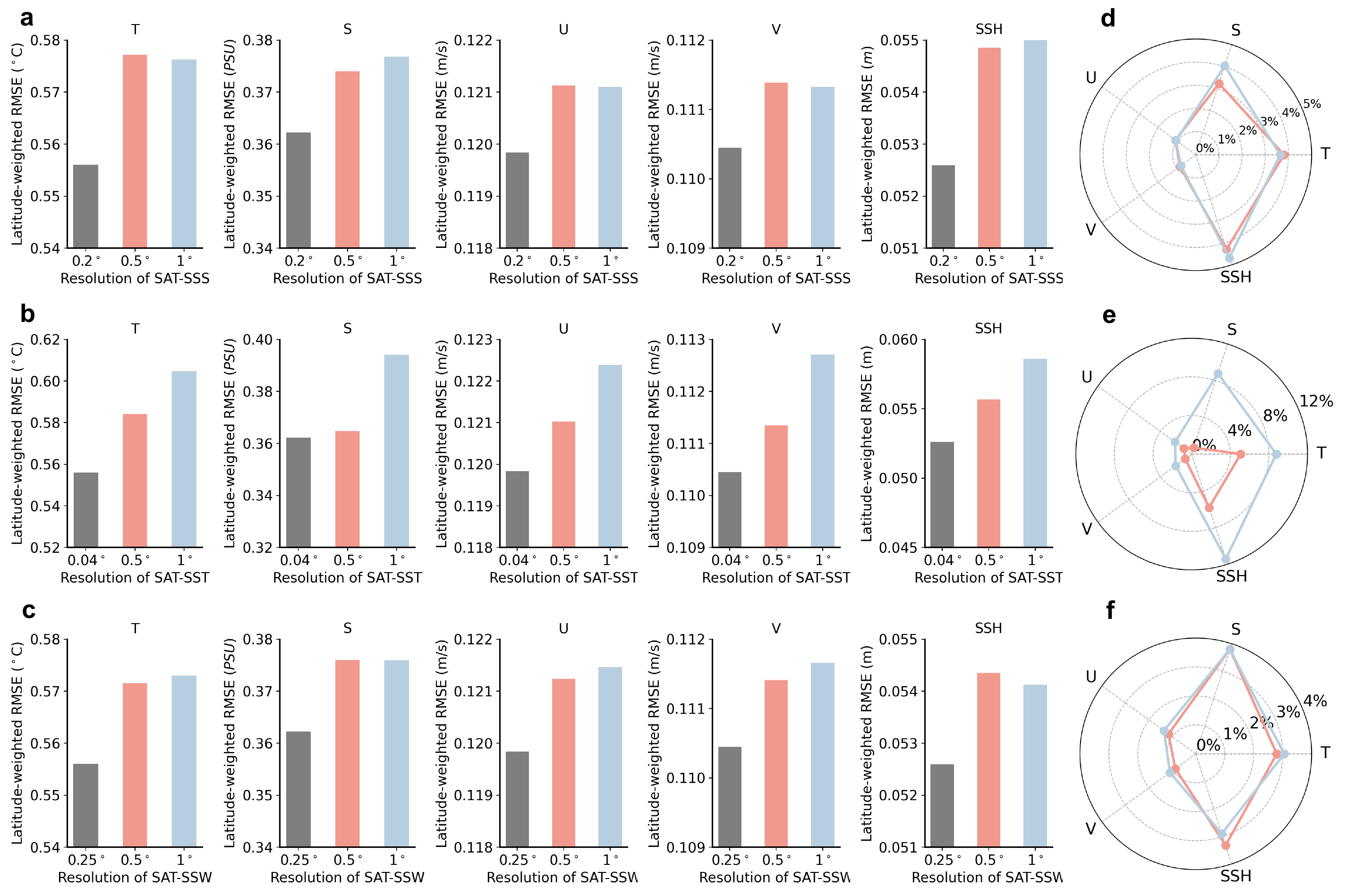}
    \caption{\textbf{Impact of satellite observation resolution on HR-DA.} \textbf{a-c} RMSE for five variables (T, S, U, V, and SSH) obtained using satellite-based (SAT) SSS, SST, and SSW observations at three resolutions: original (0.2$^\circ$ for SSS, 0.04$^\circ$ for SST, and 0.25$^\circ$ for SSW), 0.5$^\circ$, and 1$^\circ$. \textbf{d-f} Latitude-weighted RMSE reduction for each variable, comparing the original resolution to 0.5$^\circ$ and 1$^\circ$.}
    \label{appfig:diff_res}
\end{figure}

High-resolution SST observations (0.04$^\circ$) significantly improve model accuracy for T, S, and SSH (Fig.~\ref{appfig:diff_res}{a,d}). However, as SST resolution becomes coarser, the accuracy declines sharply, especially for these three variables, which are highly sensitive to SST resolution. This underscores the critical role of high-resolution SST data in improving surface ocean state estimation. 
The resolution of satellite-based SSS observations has minimal impact on the accuracy of most variables. Only slight improvements are observed for S (Fig.~\ref{appfig:diff_res}{b,e}), likely because salinity gradients are less pronounced and less variable than temperature gradients in the ocean.
For satellite-based SSW observations, interestingly, the RMSE for SSH estimation is highest at the intermediate resolution (0.5$^\circ$) but slightly lower at the coarsest resolution (1$^\circ$) (Fig.~\ref{appfig:diff_res}{c}). This suggests that higher-resolution SSW data may introduce noise or capture small-scale wind patterns that have limited influence on large-scale ocean dynamics. Fig.~\ref{appfig:diff_res}{d-f} further illustrate that SST resolution has the most significant impact on improving the accuracy of multiple variables. These findings emphasize the need for continued investment in high-resolution satellite missions to enhance ocean and climate monitoring capabilities.

\section{DL forecast model}\label{appsec:forecast_model}

In this study, we employ Triton~\cite{Wu2025Triton} as the DL forecast model, which features a hierarchical architecture designed to process information across multiple spatial and temporal resolutions, mitigating spectral bias and explicitly capturing cross-scale dynamics. Triton has demonstrated superior performance in various challenging forecasting tasks.

For training, we use the GLORYS reanalysis dataset, covering the period from 2010 to 2020. Training is conducted on data from 2010 to 2016, while data from 2017 are reserved for validation. Forecasts are performed for the independent period from 2018 to 2020, with a forecast step of one day. Due to computational constraints, the model is trained at a spatial resolution of 1$^\circ$. During inference, Triton operates in an auto-regressive manner, where the initial conditions can be sourced either from reanalysis fields or from the outputs of the DA process.

\section{Observations}

\begin{figure}[h]
    \centering
    \includegraphics[width=1\linewidth]{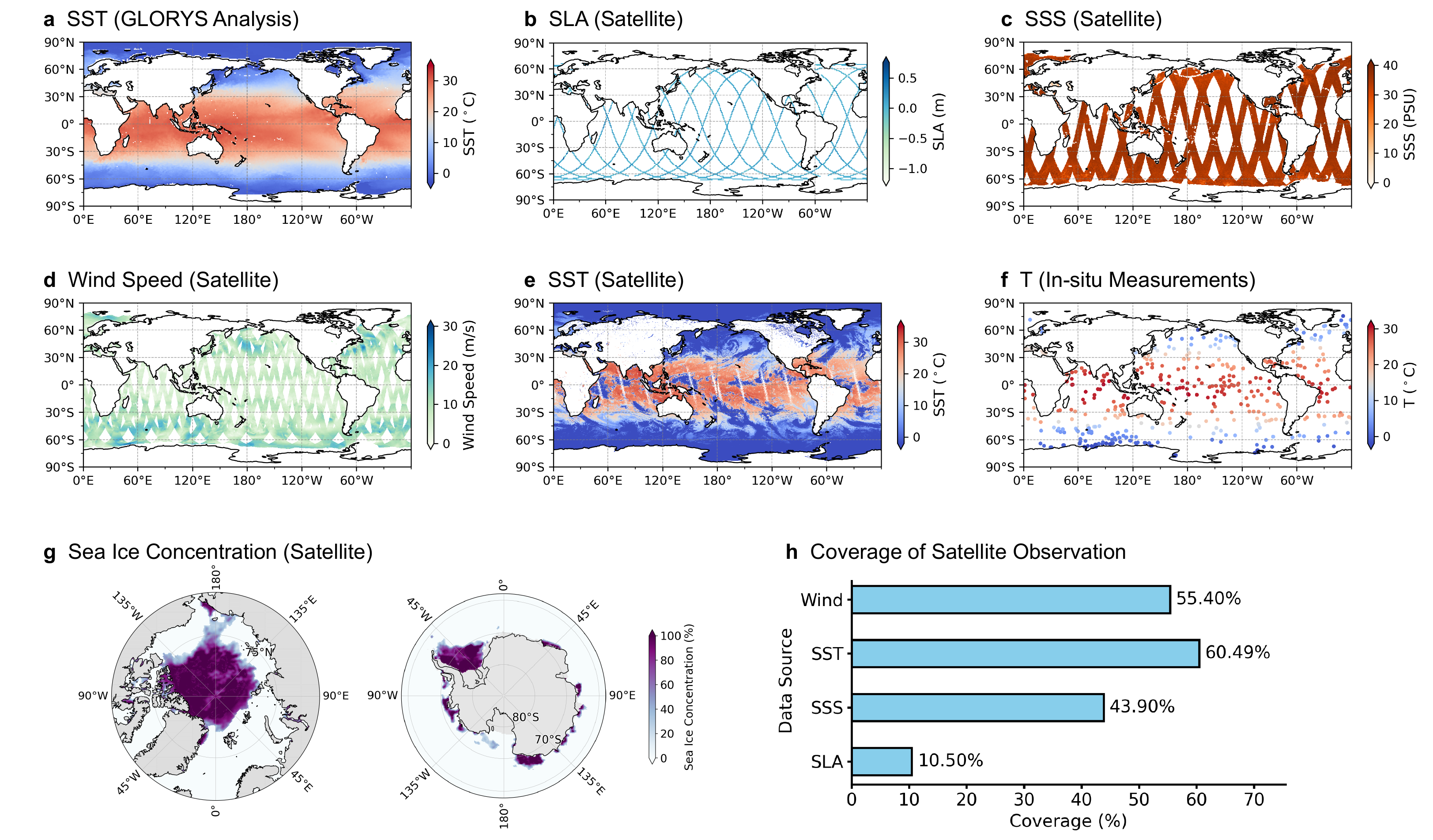}
    \caption{\textbf{Visualization of data used in this study and the coverage of satellite observations.} \textbf{a} Global Sea Surface Temperature (SST) field from the GLORYS reanalysis.\textbf{b-f} Spatial distribution of observations, including Along-track SLA, SSS, Wind Speed, SST and sparse in-situ Temperature (T) measurements. \textbf{g} Spatial distribution of SIC for the Arctic (left) and Antarctic (right). \textbf{h} Global percentage coverage of four satellite observation types (Wind, SST, SSS, SLA).}
    \label{appfig:obs_vis}
\end{figure}

\end{appendices}

\end{document}